\documentclass[11pt,a4paper]{article}
\usepackage[hyperref]{emnlp-ijcnlp-2019}
\usepackage{times}
\usepackage{latexsym}
\usepackage{graphicx}
\usepackage{booktabs}
\usepackage{epsfig}
\usepackage{amsmath}
\usepackage{amssymb}
\usepackage{listings}
\usepackage{enumerate}
\usepackage{booktabs}
\usepackage{float}
\usepackage{natbib}
\usepackage{url}

\aclfinalcopy 

\title{High Quality Real-Time Structured Debate Generation}

\author{Eric Bolton \\
  Columbia University \\
  {\tt edb2129@columbia.edu} \\\And
  Alex Calderwood \\
  Columbia University \\ 
  {\tt adc2181@columbia.edu}\\ \And
  Niles Christensen \\
  Columbia University \\
  {\tt nnc2117@columbia.edu} \\\AND
  Jerome Kafrouni \\
  Columbia University \\ 
  {\tt jk4100@columbia.edu}\\ \\\And
  Iddo Drori \\
  Columbia University \\ 
  {\tt idrori@cs.columbia.edu}\\}
  
\begin{document}

\maketitle
\begin{abstract}

Automatically generating debates is a challenging task that requires an understanding of arguments and how to negate or support them. In this work we define debate trees and paths for generating debates while enforcing a high level structure and grammar. We leverage a large corpus of tree-structured debates that have metadata associated with each argument. We develop a framework for generating plausible debates which is agnostic to the sentence embedding model. Our results demonstrate the ability to generate debates in real-time on complex topics at a quality that is close to humans, as evaluated by the style, content, and strategy metrics used for judging competitive human debates. In the spirit of reproducible research we make our data, models, and code publicly available.
\end{abstract}

\vspace{10pt}


\section{Introduction}
\label{sec:introduction}

Automatic generation of human-level debates has important applications: providing coherent deconstructions of given topics, assisting in decision-making, and surfacing previously overlooked viewpoints as demonstrated in Table \ref{table:good_examples}. Debate generation would enable policymakers and writers to improve their rhetorical abilities by practicing against an automated agent in real-time.

\begin{table}[ht]
\begin{center}
\begin{tabular}{lp{5cm}} \toprule
\textbf{Source:} & Income inequality makes people unhappy. Inequality is bad for everyone.
 \\
\textbf{Response:} & The economic outcome of an increase in income is not a good factor in determining the effects of starvation and poverty.\\
\midrule
 \textbf{Source:} & Many religious teachings promote violence. \\
\textbf{Response:} & Religious protesters have a strong history of engaging in violence against violence. Many of the people who oppose the violence are often on a case-by-case basis. \\
\bottomrule
\end{tabular}
\caption{Sample results: novel responses generated to be contradictory.}
\label{table:good_examples}
\end{center}
\end{table}

The field of debate is closely related to linguistic negation, which has studied logical rules for forming various types of negation \citep{stanfordnegation}, at the low level of words and sentence. In this work we use a large dataset and pre-trained neural language model to synthesize long debates with a high-level linguistic structure. Our work focuses on informal logic, the study of arguments in human discourse, rather than in rigorously specified formal languages \cite{stanfordinformal}.

Machine debate generation must meet several requirements: the need to respect long-distance dependencies while maintaining semantic coherence with the prompt, the need for consistent themes and tone over long durations, and finally, the need for adherence to grammar rules.

These requirements entail compelling, thoughtful responses to a given writing prompt. Models that excel at this task must understand the low-level grammar of the language, but also need to incorporate an understanding of the relationships between sentences. Thus, existing story generation models constitute a natural tool for generating arguments with an understanding of what it means to refute or support an argument's point.

\subsection{Kialo}

We collected a data set from Kialo \citep{kialo}, an online debate platform, and trained a text generation model \citep{FanLD18} to generate arguments in favor or against given debate prompts. 
Users submit responses (pros and cons), which are then approved by a moderator. Other users can then respond to those responses as though they themselves were prompts, forming a debate tree.

We leveraged the tree structure of the Kialo dataset along with its existing pro/con annotations to develop a novel self-supervised corpus creation strategy, using the insight that the metadata constraints we used to define an argument (our generative unit) effectively enforced a high-order grammar on the output. We experiment with this type of tree traversal strategy to model multi-turn text generation. This strategy is applicable to other seq2seq models that model long-form text generation.

\subsection{Contributions}

In this we work, we describe a system which learns to participate in traditional debates by modeling different sequences of argument nodes through tree-structured debates to generate prompts and responses \(corresponding to sources and targets in the translation paradigm\).

Our main contributions are:
\begin{itemize}
\item \textbf{Defining debate tree and paths for generating text:} Enforcing a high level structure and grammar, defining complex arguments, multi-turn debates, as described in Section \ref{sec:methods-debate-tree} and \ref{sec:methods-debate-paths}.
\item \textbf{Multi-turn debate generation:} An extension of the prompt to response generation paradigm to enable multi-turn generation, as described in Section \ref{sec:methods-debate-model}.
\item \textbf{Measure of debate quality:} as evaluated with criteria used in debate competitions: content, style, and strategy, repeated over multiple rounds.
\end{itemize}
Based on these key contributions, our results demonstrate near-human quality debates as shown in Section \ref{sec:results}.


\subsection{Related Work}
Our goal is to create debates that perform at a high-level according to the same evaluation metrics as human debates. Quantitatively assessing the quality of debates and arguments is a challenging task. Recent work evaluates dialogue \citep{kulikov2018importance} and stories \citep{purdy2018} by humans using Amazon's Mechanical Turk (AMT). We also use AMT for human evaluation of debates, based on the criteria commonly used in debate competitions \citep{nationaldebate}.

Our focus is debate generation, which is structured in multiple rounds. The state of the art in structured text generation focuses on stories rather than debates, and is based on generating sequences of predicates and arguments, replacing entities with names and references \citep{fan2019strategies}. Other structuring strategies include using outlines \citep{drissi2018hierarchical}, plans \citep{yao2018plan}, or skeletons \citep{skeleton2018}. All of these works build upon significant advances on the problem of text generation using hierarchical models \citep{FanLD18}. These methods can be improved and extended by using very large language models which have shown excellent results in language generation and reading comprehension \citep{devlin2018bert,peters2018deep,radford2018improving, radford2019language}.





The performance of these neural language models has relied on curating large datasets for different tasks, such as question answering \citep{squad2016} and reading comprehension \citep{qanet2018, clark2017readingcomprehension}. In contrast, our work creates a large dataset specifically for the task of generating debates. IBM's Project Debater \citep{ibmprojectdebater} also shares the same focus, however they are driven by speech and do not take advantage of debate specific datasets. There are a number of online debate hosting platforms that contain pro/con metadata \citep{procon, idebate}. We use Kialo, which uniquely combines this metadata with a tree structure that models a long multi-turn debate, depending on the parsing strategy.

\section{Methods}
\label{sec:methods}

\begin{figure*}[t]
    \centering
    \includegraphics[width=0.75\textwidth]{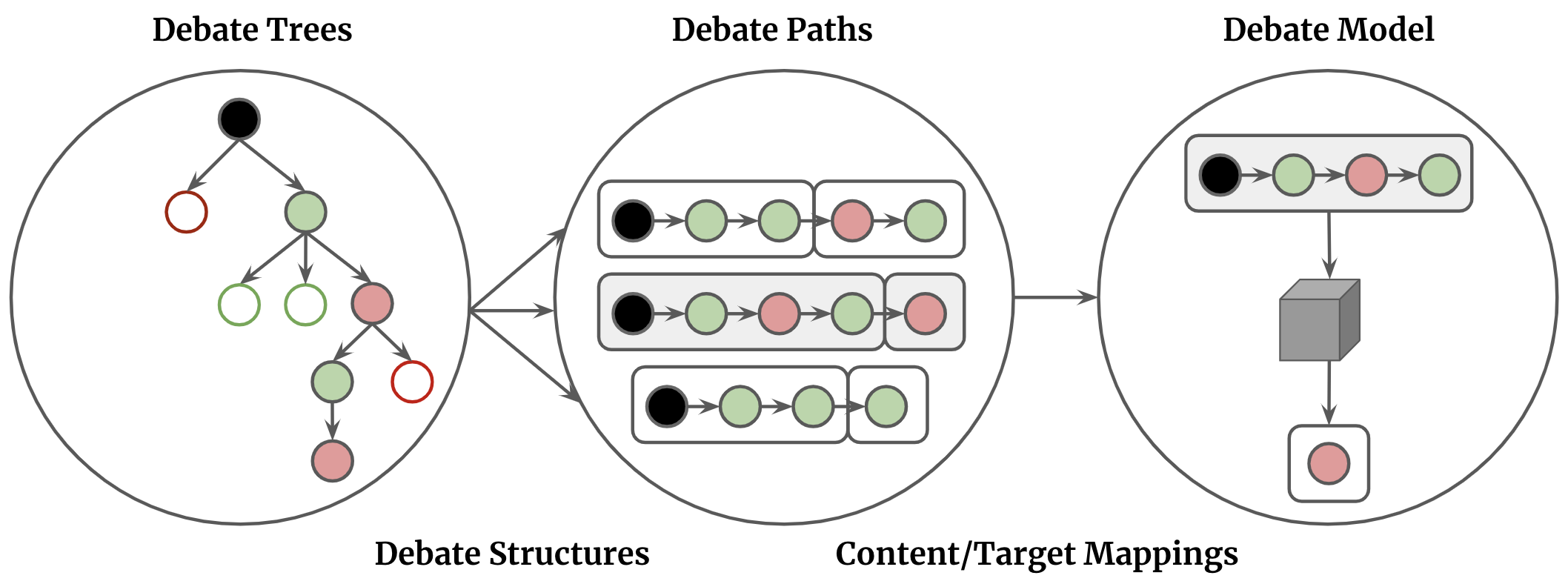}
    \caption{Our debate generation framework. We begin with debate trees on the far left, from which we extract all paths through the tree which correspond to meaningful debates, creating a dataset in the central bubble. On the far right, this new dataset is used to retrain an existing model to debate. In this paper, the retrained model is the one developed by \citet{FanLD18}. In our diagram, each green bubble supports the point above it, while each red bubble argues against the point above it.}
    \label{fig:framework}
\end{figure*}

We define a framework for generating debates from a tree-structured corpus, as illustrated in Figure \ref{fig:framework}: (i) define and collect debate tree corpus, as described in Section \ref{sec:methods-debate-tree}; (ii) build debate paths, which are content/target pairs, as described in Section \ref{sec:methods-debate-paths}; (iii) train the debate encoding on all generated debate paths, as described in Section \ref{sec:methods-debate-model}.


\begin{table*}[ht]
\small
\begin{center}
    \begin{tabular}{lrr}
        \toprule
        \textbf{Parsing Strategy} & \textbf{Prompt} & \textbf{Response} \\
        \hline
        Supportive Arguments &[Pro$|$Con][Pro]*&[Pro]+\\
        Contradicting Arguments &[Pro$|$Con][Pro]*&[Con][Pro]*\\
        Complex Arguments &[Pro$|$Con][Pro]*&[Pro$|$Con][Pro]*\\
        Multi-Turn Debates &
       [Pro$|$Con][Pro]*([Con][Pro]*)* &[Con][Pro]*\\
       \bottomrule
    \end{tabular}
\end{center}
\caption{Description of debate structures, expressed through regex notation.}
\label{table:strategies}
\end{table*}

\subsection{Debate Trees}
\label{sec:methods-debate-tree}

\begin{figure}
    \centering
    \includegraphics[width=0.5\textwidth]{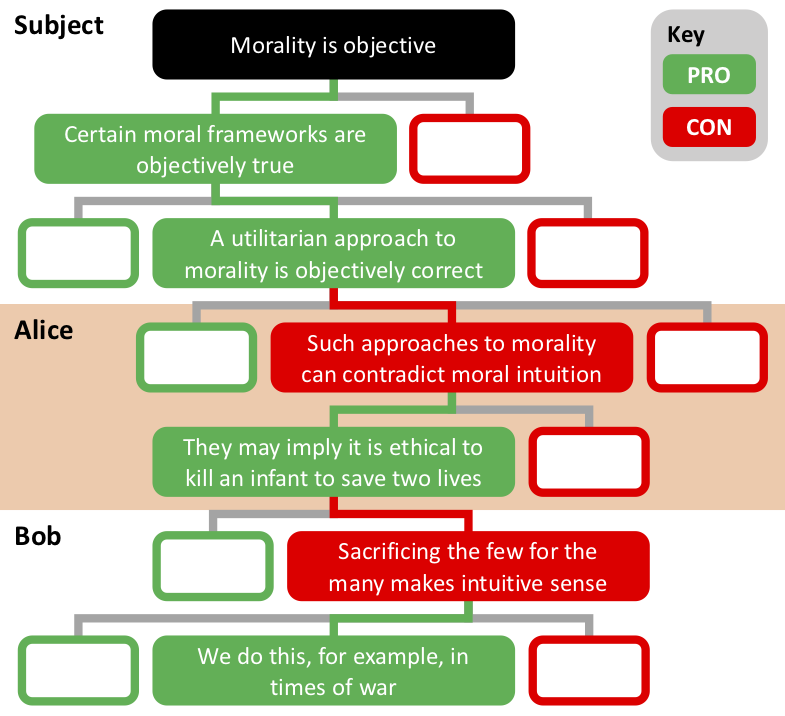}
    \caption{An example of a path through a Kialo debate tree. Green boxes indicate arguments that support their parents (Pro), while red boxes indicate arguments that contradict their parents (Con).}
    \label{fig:debate-tree}
\end{figure}

A debate tree is a tree with one extra bit of storage for each node, its stance: either `Pro' or `Con' depending on whether the node supports or refutes its parent argument. A debate is a path on this debate tree, for example from the root to a leaf, as shown in Figure \ref{fig:debate-tree}. An argument is defined as one node in tree.

We extract debate tree structures from a public debate site \citep{kialo} which allows crawling, and filter for English debates \cite{langdetect}. When Kialo users referenced different parts of the debate tree, which rarely happened, we copy the text in the referenced node into the referring node.


\begin{table*}[ht] 
\begin{center}
\small
\begin{tabular}{lrrrr}
    \toprule
     & \textbf{Complex} & \textbf{Supportive} & \textbf{Contradictory} & \textbf{Multi-Turn} \\  \hline
    \# Training examples & 432,999 & 222,852 & 211,355 & 459,643 \\
    \# Test examples & 45,471 & 21,946 & 17,677 & 30,607 \\
    \# Validation examples & 38,779 & 20,810 & 14,485 & 35,916 \\
    \hline
     Prompt dictionary size &  20,535 & 14,691 & 16,690 & 50,626 \\
     Response dictionary size &  28,313  & 21,907 & 19,534 & 55,359 \\
     \bottomrule
\end{tabular}
\caption{Description of datasets created from \textsc{Kialo} using each tree-traversal strategy.}
\label{table:statistics}
\end{center}
\end{table*}


\subsection{Debate Paths}
\label{sec:methods-debate-paths}
The debate tree can be parsed in different ways depending on the desired properties of the resulting corpus. We present four parsing strategies. For each, we provide a regex-like description corresponding to the prompt and the response of the debate, as described in Table \ref{table:strategies}.

The majority of paths through the tree would correspond to nonsensical debates. In order to model sensible and internally coherent debates in which each modeled participant has a clear viewpoint, we developed the following parsing strategies:

\paragraph{Supportive Arguments} - a prompt and a response supporting it.
\paragraph{Contradicting Arguments} - a prompt and a response refuting it.
\paragraph{Complex Arguments} - a prompt and a response that may either support the prompt or refute it. This is the union of all arguments generated by the supportive and contradicting arguments approaches.


\paragraph{Multi-Turn Debates} - a prompt containing a back-and forth debate between two speakers, separated by a specific token. Each participant is constrained to always disagree with the previous statement made by their opponent. This approach allows for the generation of debates of arbitrary length.



\subsection{Debate Model}
\label{sec:methods-debate-model}

Once the debate paths are extracted, they are used to train a text generation model. In this work we retrained the model developed by \citet{FanLD18}, but another model, such as the one developed by \citet{devlin2018bert}, could be substituted in its place with minimal changes.

To determine what would yield the best generation performance, we trained models with our complex arguments dataset under various conditions. We filtered for only Pro responses and Con responses, and compared results between Fusion \citep{SriramJSC17} and plain seq2seq models \citep{gehring2017convolutional}. 


We limited our vocabulary to tokens appearing 10 or more times in the training set and replaced words that don't fit this criterion with an unknown token. We used named entity recognition (NER) \citep{spacy2} to limit the number of unknown tokens on our complex arguments datasets, to improve model coherence. Finally, we modeled an end of argument token to delimit arguments. We used 5\% of the debates for our validation set, and 5\% for our test set. A detailed description of our resulting dataset is shown in Tables \ref{table:statistics}. We implemented a custom generation approach to use previously generated responses as input prompts for real-time generation of new responses. This enabled generation of multi-turn debates using the seq2seq model trained on our debate data.

\section{Results}

\begin{table*}[t]
\small
\begin{center}
\begin{tabular}{cp{13cm}}\toprule
\textbf{Subject:} & All religions should be banned on a global scale. \\
\midrule
\textbf{Generated debate:} \\
\textbf{Alice:} & Religion is a global problem that cannot be solved in a bad way. Religion is a tool used to challenge human rights. It does not exist to be as a way to be a good thing for humans in order to use the world as a form of language and the existence of its own own religion. For example, religion is not the only way to prevent a world from a different religion \\
\textbf{Bob:} &  Religion is a function in which morality involves the existence of something, rather than just the existence of one another. Religion is a personal and philosophical concept \\ 
\textbf{Alice:} &  Religion is a form of tradition. Simply because one group may be sacrificed or not. Everyone can make a person considered to be good at all \\
\textbf{Bob:} &  Religion may be a good thing for many people to commit violence, but rather the life of one person \\
\textbf{Alice:} &  Religion is not a way to measure a belief that it is inherently acceptable to kill a sandwich if the one needs it gets to live \\ 
\textbf{Bob:} &  Not all people are the ones doing anything of anything. Religion is the gift of that they disagree about. This is the same as the people who value the community may value their lives through little reason not to process it or destroy people or other people to live \\
\textbf{Alice:}  &  Moral values are a way to value moral and ethical consequences, no one simply wants to do what is right, not worth how you want \\
\textbf{Bob:} &  If people have a duty to help those people to value their life they will have to act in the interest of their place rather than forcing them to do so without harming their religious life \\
\bottomrule
\end{tabular}
\caption{Multi-turn debate generated from human prompt ``All religions should be banned on a global scale.''}
\label{table:long_results}
\end{center}
\end{table*}

\label{sec:results}

Our results demonstrate the ability to generate debates in real-time on complex topics, as shown in Table \ref{table:long_results}. Evaluating complex debates is a difficult task. Perplexity evaluation, shown in Table \ref{table:perplexity}, is limited, and does not capture debate quality well. Therefore we use human evaluation of our generated debates \cite{kulikov2018importance}. Specifically, we use Amazon Mechanical Turk \citep{mturk} for evaluating multi-turn debates, with 119 distinct turkers and a total of 526 evaluations times 4 criteria on a 1-4 scale.



\begin{table}
\begin{center}
\small
\begin{tabular}{ccc} \toprule
    \textbf{Model type} & \textbf{Strategy} & \textbf{Perplexity} \\
    \hline
    seq2seq & No NER, Complex & 73.16 \\
    seq2seq & No NER, Supportive & 54.68\\
    seq2seq & No NER, Contradicting & 60.72\\
    seq2seq & NER, Complex & 80.87\\
    seq2seq & NER, Supportive & 58.82\\
    seq2seq & NER, Contradicting & 77.39\\
    Fusion & No NER, Complex & 70.41\\
    Fusion & No NER, Supportive & 43.06\\
    Fusion & No NER, Contradicting & 60.33\\
    Fusion & NER, Complex & 83.77 \\
    Fusion & NER, Supportive & 77.04\\
    Fusion & NER, Contradicting & 79.98\\
    \bottomrule
\end{tabular}
\caption{Average perplexity over all generated debates.}
\label{table:perplexity}
\end{center}
\end{table}

We compare our generated debates with human debates, by comparing debate paths from the test set with debate sequences predicted by our model. We generated lengthy debates of between 5-15 back-and-forths and evaluated debates of length 10, with human debates of the same length, preventing any bias from the Turkers for longer or shorter debates. We asked humans to rank 50 human-created and 56 automatically-generated debates, where each debate was rated by 5-10 separate reviewers for a total of 2,104 answers as shown in Table \ref{table:mturk}. Reviewers were asked to evaluate the style, content, strategy, and overall quality of each debate, which are the same categories used to judge competitive debaters in the US National Debate Tournament \citep{nationaldebate}. Our results demonstrate generated debates which are evaluated to be close to human level performance as shown in Table \ref{table:mturk}.

\subsection{Implementation}


Our model was trained and run on a virtual machine running Ubuntu 16.04 on 8 vCPUs and 2 NVIDIA Tesla K80 GPUs. The machine had access to 52 GB of memory and 100 GB of storage.
Training time is 3 days on this machine. Given a new subject for debate, average computation time of a new argument is around 3 seconds, which is acceptable in real-time debates.  In the spirit of reproducible research we make our data, models, and code available \citep{ourcode}.


\begin{table}
\begin{center}
\small
\begin{tabular}{ccccc}
\toprule
\textbf{Criteria} & \textbf{Human} & \textbf{Generated}\\
\hline
Style  &  $3.273 \pm 0.722$ & $3.154 \pm  0.868$ \\
\hline
Content & $3.392 \pm 0.626$  &  $3.289 \pm  0.770$ \\
\hline
Strategy & $3.412 \pm 0.596$ & $3.133 \pm  0.821$ \\
\hline
Overall & $3.512 \pm  0.596$ & $3.193 \pm 0.824$ \\
\bottomrule
\end{tabular}
\caption{Mean and standard deviation of Mechanical Turk evaluations. Ratings were on a 1-4 scale.}
\label{table:mturk}
\end{center}
\end{table}



\subsection{Limitations}

Unknown tokens present in the output were replaced with underscores representing an underline, and turkers were shown a note explaining that certain words would be omitted and replaced with an underline. Therefore, these missing words may have skewed the Turkers' judgments on responses unknown tokens, such as in Table \ref{table:bad_examples}. 

\begin{table}
\begin{center}
\small
\begin{tabular}{lp{5cm}} \toprule
\textbf{Source:} & People have been led to do terrible things in the name of religion. \\
\textbf{Response:} & Many religions have the option of {\ttfamily <unk>} \\
\bottomrule
\end{tabular}
\caption{A generated response that suffered from a lack of coherence due to unknown tokens.}
\label{table:bad_examples}
\end{center}
\end{table}

Our generated debates are extremely varied, ranging from short simple quips to complicated multi-sentence responses. This results in a high standard deviation of perplexities for each debate we generate. Whether or not the debate was complete (addressing important points) was not a criterion that we assessed.

\section{Conclusion and Future Work}

This work builds a dataset of varied arguments and responses from a tree-structured debate corpus by leveraging information about the nature of the arguments. This approach resulted in a large dataset of varied debate structures.








We present and evaluate a novel structured debate generation approach which brings us closer to realistic simulation of human debate. Our contribution is a multi-turn debate structure following a high level stance-based grammar. 




Our framework is language model agnostic; in the future we would like to improve our work using large embeddings to experiment with this and other text generation models.

\bibliography{paper}
\bibliographystyle{acl_natbib}

\end{document}